\documentclass{article}

\usepackage{microtype}
\usepackage{graphicx}
\graphicspath{{figures/}}
\usepackage{subcaption}
\usepackage{booktabs} 
\usepackage{makecell} 

\usepackage{hyperref}

\usepackage[accepted]{icml2026}

\usepackage{amsmath}
\usepackage{amssymb}
\usepackage{mathtools}
\usepackage{amsthm}

\usepackage{algorithm}
\usepackage{algorithmic}

\usepackage[capitalize,noabbrev]{cleveref}

\theoremstyle{plain}

\theoremstyle{definition}

\theoremstyle{remark}

\usepackage[textsize=tiny]{todonotes}

\icmltitlerunning{Self-Mined Hardness for Safety Fine-Tuning}

\begin{document}

\twocolumn[
  \icmltitle{Self-Mined Hardness for Safety Fine-Tuning}

  \icmlsetsymbol{equal}{*}

  \begin{icmlauthorlist}
    \icmlauthor{Prakhar Gupta}{umich}
    \icmlauthor{Garv Shah}{umich}
    \icmlauthor{Donghua Zhang}{umich}
  \end{icmlauthorlist}

  \icmlaffiliation{umich}{University of Michigan}

  \icmlcorrespondingauthor{Prakhar Gupta}{prakharg@umich.edu}

  \icmlkeywords{jailbreak, safety, fine-tuning, hard-example mining, LoRA, overrefusal}

  \vskip 0.3in
]


\printAffiliationsAndNotice{}  

\begin{abstract}
  Safety fine-tuning of language models typically requires a curated
  adversarial dataset. We take a different approach: score each
  candidate prompt's difficulty by how often the target model's own
  rollouts are judged harmful, then fine-tune on the hardest prompts
  paired with the model's own non-jailbroken rollouts. On Llama-3-8B-Instruct and
  Llama-3.2-3B-Instruct, this approach cuts the WildJailbreak attack
  success rate from $11.5\%$ and $20.1\%$ down to $1\text{--}3\%$,
  but pushes refusal on jailbreak-shaped benign prompts from
  $14\text{--}22\%$ to $74\text{--}94\%$. Interleaving the same hard
  prompts $1{:}1$ with adversarially-framed benign prompts (prompts
  that look like jailbreaks but have benign intent) cuts that
  refusal back down to $30\text{--}51\%$ on 8B and $52\text{--}72\%$
  on 3B, at a cost of $2\text{--}6$ percentage points of attack
  success rate. Within the mixed regime, training on the hardest
  half of the eligible pool rather than a random half cuts the
  remaining ASR by $35\text{--}50\%$ (about $3$ percentage points)
  on both models.
\end{abstract}

\section{Introduction}

Alignment fine-tuning \citep{ouyang2022instructgpt,bai2022constitutional}
makes instruction-tuned language models less likely to produce
harmful content. But adversarial prompts (role-play setups,
hypothetical framings, persona instructions) can get them to
answer harmful requests they would otherwise refuse
\citep{wei2023jailbroken,zou2023gcg}. The standard fix is more
safety-tuning on adversarial prompts, drawn either from human attack
corpora like WildJailbreak \citep{jiang2024wildteaming} or from
hand-designed clean/biased pairs like those in Bias-Augmented
Consistency Training \citep{chua2024bct}. Both approaches start
from a curated adversarial dataset.

We take a different approach. For each candidate adversarial prompt,
we sample $K$ rollouts (responses) from the target model and label
each rollout with a three-judge safety ensemble. A prompt's
\emph{harmful rate}, the fraction of rollouts majority-labeled as
harmful, becomes its difficulty score. Prompts
that are sometimes but not always jailbroken form the \emph{eligible
pool}: these have both a training signal to learn from and a
non-jailbroken rollout we can use as the supervised target. Sorting
this pool by harmful rate gives a natural ordering from hardest to
easiest. We need no external clean dataset or hand-designed
adversarial families.

But this approach has a clear tradeoff. Training on the hardest
self-mined prompts drives WildJailbreak ASR from $11.5\%$ down to
$2\text{--}3\%$ on Llama-3-8B-Instruct and from $20.1\%$ down to
about $1\%$ on Llama-3.2-3B-Instruct \citep{dubey2024llama3}, but
pushes refusal on plain benign
prompts from about $5\%$ to $32\text{--}55\%$, and refusal on
jailbreak-shaped benign prompts from $14\text{--}22\%$ to
$74\text{--}94\%$ (\S\ref{sec:results-pure}). The model has learned
to refuse anything that looks like a jailbreak, not just
harmful requests. We interleave the adversarial sequence $1{:}1$
with adversarially-framed benign
prompts (\emph{adv-benign} for short): long, role-play-style
prompts that look like jailbreaks but have benign intent, paired
with the model's non-refused rollouts.
Mixing brings refusal on jailbreak-shaped benign prompts down to
$30\text{--}51\%$ on 8B and $52\text{--}72\%$ on 3B, at a cost of
$2\text{--}6$ percentage points of ASR. The adversarial half is identical to the pure version.

We compare five fine-tuning baselines (\emph{Hard}, \emph{Random},
\emph{Control}, \emph{Hard-Mixed}, \emph{Random-Mixed}) and report
each at its $50\%$-of-pool checkpoint. At that checkpoint, Hard has
trained on the top $50\%$ hardest prompts from the eligible pool,
Random on a random $50\%$ of that same pool, and Control on the same
number of vanilla benign prompts. Hard-Mixed and Random-Mixed
additionally interleave the Hard or Random adversarial sequence
$1{:}1$ with adversarially-framed benign prompts. All five baselines
share a canonical (prompt $\to$ safe-response) pairing, so any
difference between Hard and Random is due only to which half of the
pool the checkpoint has seen. The same $50\%$ rule is applied to
each model separately. Past that point, the remaining prompts are
too easy to add useful adversarial signal
(\S\ref{sec:cutoff}).\footnote{Code:
\url{https://github.com/prakharg55/jailbreak-ICML-FA}}

\paragraph{Contributions.}
\begin{itemize}
  \item A self-contained pipeline that mines per-prompt jailbreak
        difficulty from the target model's own rollouts and pairs the
        hardest prompts with non-jailbroken rollouts as supervised
        training data.
  \item A $1{:}1$ interleaving with adversarially-framed benign
        prompts as a targeted fix for the overrefusal that pure
        adversarial fine-tuning causes, evaluated against pure and
        random-selection baselines under a canonical prompt-to-target
        pairing.
  \item Test-set results on Llama-3-8B-Instruct and
        Llama-3.2-3B-Instruct across three safety and three
        overrefusal benchmarks, showing the tradeoff and the
        mitigation on both models.
\end{itemize}

\section{Related Work}

\paragraph{Safety alignment.} Instruction tuning with human-preference
data \citep{ouyang2022instructgpt} and AI-feedback variants
\citep{bai2022constitutional} are the standard approach for reducing
a model's propensity to comply with harmful requests. Open-weights
instruction-tuned models such as Llama~3 \citep{dubey2024llama3} ship
with this training already applied.

\paragraph{Jailbreak attacks and benchmarks.} Adversarial prompting
\citep{wei2023jailbroken} and automated attack search
\citep{zou2023gcg} reliably elicit harmful completions from aligned
models. WildJailbreak and WildGuardMix
\citep{jiang2024wildteaming,han2024wildguard} contain large pools of
adversarial prompts with judged responses. ClearHarm
\citep{kuhn2024clearharm} provides clearly harmful prompts.

\paragraph{Consistency training.} Bias-Augmented Consistency Training
(BCT) \citep{chua2024bct} trains models on paired
\emph{clean}/\emph{biased} prompts to preserve clean behavior under
adversarial reformulations. \citet{irpan2025consistency}
concurrently apply BCT directly to jailbreaks and sycophancy,
using the model's own responses as training targets. Our design is
motivated by BCT but differs on two axes: the adversarial
distribution is selected by self-mined per-prompt hardness (rather
than paired prompt variants), and we include a prompt-selection
ablation (hardest-half vs random-half) that a consistency-only
objective does not isolate.

\paragraph{Curriculum learning and hard-example mining.} Presenting
examples in a meaningful order can affect optimization dynamics
\citep{bengio2009curriculum}. In computer vision, online hard-example
mining focuses gradient updates on the currently misclassified
examples \citep{shrivastava2016ohem}. We adapt this idea to safety
fine-tuning at the prompt level: each prompt's difficulty is estimated
by how often the model's own rollouts on it are judged harmful. We use
LoRA \citep{hu2022lora} adapters as the parameter-efficient
fine-tuning backbone.

\section{Method}

\subsection{Notation}

Let $M$ be a fixed instruction-tuned target model. Given an adversarial
prompt $p$ we draw rollouts $r \sim M(\cdot \mid p)$ at sampling
temperature $T = 1$. A safety judge maps a (prompt, response) pair to
\textsc{harmful} or \textsc{unharmful}. With three judges $J_1, J_2,
J_3$, the majority label $\overline{J}(p, r)$ is \textsc{harmful} when
at least two judges return \textsc{harmful} (and analogously for
\textsc{unharmful}). For $K$ rollouts $r_1, \dots, r_K$ on prompt $p$,
the \emph{harmful rate} is
\begin{equation}
  \mathrm{hr}(p) = \tfrac{1}{K}\,\bigl|\{i \mid \overline{J}(p, r_i) =
    \textsc{harmful}\}\bigr|.
\end{equation}
We call $p$ \emph{eligible} if $0 < \mathrm{hr}(p) < 1$: only such
prompts admit both a supervised target (a non-jailbroken rollout) and
a meaningful learning signal (some failures to learn from).

\subsection{Hard-Example Mining}
\label{sec:mining}

We sample adversarial prompts $\mathcal{P}_h$ from the
\texttt{adversarial\_harmful} partition of WildJailbreak
\citep{jiang2024wildteaming}, generate $K_h = 64$ rollouts per prompt,
and classify each rollout with WildGuard \citep{han2024wildguard},
MD-Judge v0.1 \citep{li2024saladbench}, and Llama-Guard-3
\citep{inan2023llamaguard}. Three judges with distinct templates make
the majority vote less sensitive to the idiosyncratic errors any one
classifier makes on the long, role-play-style prompts that dominate
WildJailbreak. The eligible set
$\mathcal{E} = \{p : 0 < \mathrm{hr}(p) < 1\}$ feeds into every
downstream baseline. For each eligible prompt, we pick one of its
non-jailbroken rollouts uniformly at random as the training target,
and reuse that same (prompt, target) mapping across every baseline,
so any difference between Hard and Random comes only from which
prompts each has seen, not from different training responses.

\subsection{Adversarially-Framed Benign Supervision}
\label{sec:advbenign}

Overrefusal is the opposite failure mode of safety tuning. It is
driven by prompts that look like jailbreaks (role-play
preambles, hypothetical framings, persona instructions) but whose
underlying request is harmless. WildJailbreak's
\texttt{adversarial\_benign} partition contains exactly this kind
of prompt. To teach the model to tell apart looks like a jailbreak from actually asks for harm, we mine non-refused
rollouts for these prompts.

For each prompt $q$ sampled from the adv-benign pool, we draw
$K_b = 4$ rollouts at $T = 1$ and classify each for refusal using
WildGuard \citep{han2024wildguard}. We retain $q$ if at least one of
the four rollouts is non-refused and pick one of those as the
compliant training target (again, the same target is reused across
baselines). We use $K_b = 4$ because we only need one non-refused
rollout per prompt. The 8B model's per-rollout refusal rate on these
prompts is $17.5\%$, so $K_b = 4$ retains $91\%$ of sampled prompts.
We apply the same $K_b$ to 3B and it yields a similar survival rate.
Refusal is also a more regular distinction than harmfulness, which
lets us use a single classifier instead of the full three-judge
ensemble.

\subsection{The Five Baselines}
\label{sec:baselines}

Let $n = |\mathcal{E}|$. We sample a size-$n$ subset $\mathcal{B}$ of
the eligible adv-benign prompts and a size-$n$ sample
$\mathcal{V}$ of vanilla benign prompts. \Cref{tab:baselines}
summarises the five resulting datasets, and \cref{alg:mining} (in the
appendix) gives the full construction procedure.

\begin{table*}[t]
  \caption{The five fine-tuning baselines. $\mathcal{E}$ is the
    eligible adversarial-harmful set; $\mathcal{B}$ is a size-$n$
    sample of eligible adv-benign prompts; $\mathcal{V}$ is a
    size-$n$ sample of vanilla benign prompts. Hard-side prompts are
    paired with their canonical safe target
    $\mathrm{safe\_target}(p)$; benign-side prompts with their
    canonical compliant target $\mathrm{compliant\_target}(q)$;
    control prompts with one $T{=}1$ rollout from the target model.}
  \label{tab:baselines}
  \vskip 0.1in
  \begin{center}
    \begin{small}
      \begin{tabular}{lllll}
        \toprule
        Baseline      & Prompts                                & Length & Order                                                & Purpose                              \\
        \midrule
        Hard          & $\mathcal{E}$                          & $n$    & $\mathrm{hr}$-descending (stable)                    & train on the hardest half of $\mathcal{E}$   \\
        Random        & $\mathcal{E}$                          & $n$    & random permutation                                   & train on a random half of $\mathcal{E}$; selection ablation for Hard \\
        Control       & $\mathcal{V}$                          & $n$    & random permutation                                   & compute-matched null                 \\
        Hard-Mixed    & $\mathcal{E} \cup \mathcal{B}$         & $2n$   & alternating; Hard in $\mathrm{hr}$-desc order        & safety $+$ overrefusal mitigation    \\
        Random-Mixed  & $\mathcal{E} \cup \mathcal{B}$         & $2n$   & alternating; Random in random order                  & selection ablation under mixing      \\
        \bottomrule
      \end{tabular}
    \end{small}
  \end{center}
  \vskip -0.1in
\end{table*}

This construction has two properties worth noting. First,
the same shared benign sequence $\mathcal{B}$ appears at the same
positions in both Hard-Mixed and Random-Mixed. Only the
adversarial half differs. Second, because the interleave is strict
$[H_0, B_0, H_1, B_1, \dots]$ with no shuffling, Hard at any pure
checkpoint $k$ and Hard-Mixed at mixed checkpoint $2k$ have trained
on exactly the same first $k$ adversarial prompts in the same
order. The pure-vs-mixed comparison is therefore isolated to the
addition of the interleaved benign half.

\subsection{Fine-Tuning and Reporting Cutoff}
\label{sec:cutoff}

We fine-tune $M$ via LoRA \citep{hu2022lora} adapters per baseline.
Each baseline's JSONL file is read top to bottom, so at any
intermediate checkpoint Hard has consumed the hardest prompts
first and Random a random subset of the same pool
(\cref{tab:baselines}). Full hyperparameters are in
\cref{tab:hyperparams}. For each model we report at the point
where the Hard baseline has trained on the top $50\%$ of its
eligible pool.

\paragraph{Why $50\%$.} The bottom half of the eligible pool is
nearly all easy prompts on both models. On 8B (pool size
$|\mathcal{E}| = 2{,}388$), of the $1{,}194$ prompts in the bottom
half, $76\%$ have $\mathrm{hr}(p) \leq 5\%$ (jailbroken at most
$3$ of $64$ rollouts) and $39\%$ sit at the empirical floor
$\mathrm{hr}(p) = 1/K_h \approx 1.6\%$ (jailbroken exactly once).
3B shows the same pattern (pool size $2{,}488$): of the $1{,}244$
bottom-half prompts, $70\%$ have $\mathrm{hr}(p) \leq 5\%$ and
$37\%$ sit at the floor. Training past this point adds prompts
with almost no adversarial signal on either model. The $50\%$
cutoff corresponds to a hardness threshold of $\tau \approx 0.094$
on 8B and $\tau \approx 0.125$ on 3B (``prompts the base model
fails on at least about $10\text{--}13\%$ of the time''). The full
harmful-rate distribution is in \cref{tab:hardness-dist}.

\section{Experimental Setup}

\subsection{Target Models}
We run the full pipeline on two target models from the Llama-3 family
\citep{dubey2024llama3}: \texttt{meta-llama/Meta-Llama-3-8B-Instruct}
(referred to as \emph{8B}) and
\texttt{meta-llama/Llama-3.2-3B-Instruct} (\emph{3B}). Both models
are instruction-tuned open-weights checkpoints. The 3B model lets us
test whether the qualitative findings carry across scale, while
sharing tokenizer and prompt format with the 8B run for an
even comparison. Inference for rollouts, judging, and
evaluation is served through vLLM \citep{kwon2023vllm}. LoRA adapters
are swapped per checkpoint without reloading the base model.

\subsection{Mining Scales}
We apply the same mining protocol to both target models. On the
hard-mining side, we sample $N_h = 5{,}000$ adversarial-harmful
prompts and draw $K_h = 64$ rollouts per prompt. \Cref{tab:yield}
reports the eligible counts: $2{,}388$ on 8B and $2{,}488$ on 3B.

On the benign side, we need exactly $|\mathcal{E}|$ adv-benign
prompts to pair $1{:}1$ with the eligible hard set. We sample
$N_b = 3{,}000$ prompts at $K_b = 4$ rollouts each (enough to buffer
against the refusal filter, which drops prompts where none of the
four rollouts is non-refused) and keep the first $|\mathcal{E}|$
that pass. At the observed $17.5\%$ per-rollout refusal rate on 8B,
$K_b = 4$ retains over $90\%$ of sampled prompts, comfortably above
both models' eligible counts, and 3B has a similar survival rate.

\begin{table}[t]
  \caption{Adversarial-harmful mining yield on both target models, from
    the same $5{,}000$-prompt sample with $K_h = 64$ rollouts each.
    ``Never jailbroken'' prompts give no hardness signal, ``always
    jailbroken'' prompts have no safe rollout to use as a target, and
    only the \emph{eligible} set enters the Hard and Random baselines. The
    smaller 3B model is more vulnerable, so fewer of its rollouts are
    refusals: it has fewer ``never jailbroken'' prompts and ends up
    with a slightly larger eligible pool.}
  \label{tab:yield}
  \vskip 0.1in
  \begin{center}
    \begin{small}
      \begin{tabular}{lrr}
        \toprule
        Metric                                                & 8B       & 3B       \\
        \midrule
        Sampled adversarial-harmful prompts                   & $5{,}000$ & $5{,}000$ \\
        Never jailbroken ($\mathrm{hr}(p) = 0$)               & $2{,}607$ & $2{,}504$ \\
        Always jailbroken ($\mathrm{ur}(p) = 0$, dropped)     & $5$       & $8$       \\
        \textbf{Eligible}                                     & $\mathbf{2{,}388}$ & $\mathbf{2{,}488}$ \\
        \bottomrule
      \end{tabular}
    \end{small}
  \end{center}
  \vskip -0.1in
\end{table}

\subsection{Judges}
Safety judging uses the majority vote of WildGuard
\citep{han2024wildguard}, MD-Judge v0.1 \citep{li2024saladbench}, and
Llama-Guard-3 \citep{inan2023llamaguard}. Refusal classification uses
WildGuard alone. All judges run with greedy decoding.

\subsection{Evaluation Datasets}
We evaluate on six held-out test sets: three safety and three
overrefusal. The safety sets are the WildJailbreak eval split, the
adversarial-harmful partition of WildGuardMix
\citep{han2024wildguard}, and ClearHarm \citep{kuhn2024clearharm}.
The overrefusal sets are \texttt{wildjailbreak\_adv\_benign}
(adversarially-framed benign prompts from WildJailbreak's eval
split), \texttt{wildguardmix\_adv\_benign} (adversarial-yet-unharmful
prompts from WildGuardMix), and \texttt{wildguardmix\_benign}
(plain benign prompts from WildGuardMix). The first two overrefusal
sets probe the failure mode we care about most (prompts that share
the surface structure of jailbreaks but have benign intent). The
third acts as a sanity check on plain benign prompts.

\paragraph{Metrics.} On safety datasets we report the attack success
rate $\mathrm{ASR}$ under three-judge majority voting. On overrefusal
datasets we report the mean refusal rate from WildGuard. All
evaluation generations use greedy decoding.

\section{Results}
\label{sec:results}

\begin{figure*}[t]
  \centering
  \includegraphics[width=\textwidth]{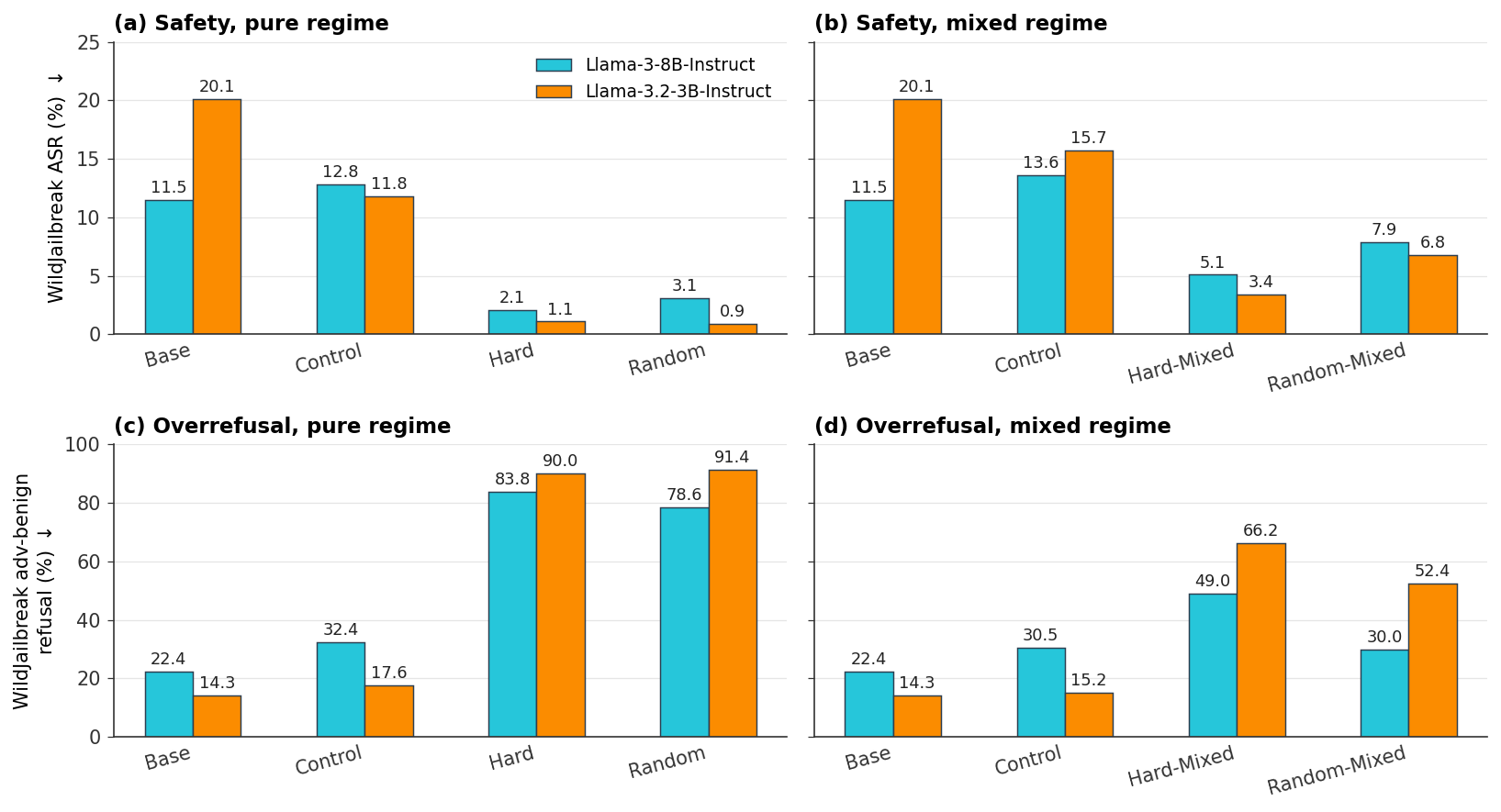}
  \caption{Headline test-set results. Lower is better in every panel.
    Each column is internally compute-matched: \textbf{left}
    (\textbf{a, c}) = pure regime (top $50\%$ of each model's
    eligible pool). \textbf{Right} (\textbf{b, d}) = mixed regime
    (the same hardest prompts interleaved $1{:}1$ with an equal-sized
    adv-benign set, so twice the training prompts). The Control
    bar in each subplot is the matched-compute reference for that
    regime: a different control checkpoint in (a, c) than in (b, d).
    \textbf{Top row (a, b):} WildJailbreak attack success rate. Pure
    Hard and Random drive ASR near zero on both models. Hard-Mixed
    and Random-Mixed pay a small ASR cost.
    \textbf{Bottom row (c, d):} WildJailbreak adv-benign refusal rate
    (the overrefusal failure mode). Pure Hard and Random push refusal
    above $78\%$ on both models. Mixing brings refusal back to
    $30\text{--}66\%$.}
  \label{fig:summary}
\end{figure*}

We report test-set results on both models at the $50\%$-of-pool
checkpoint selected in \S\ref{sec:cutoff}. Each model is evaluated
at its own checkpoint: the pure baselines (Hard, Random, Control)
train on the top $50\%$ of that model's eligible pool, and the
mixed baselines interleave that same set $1{:}1$ with an equal-sized
adversarially-framed benign set. \Cref{tab:test-results} reports
the 8B numbers and \cref{tab:test-results-3b} the 3B numbers. Both
span all three safety and three overrefusal benchmarks.
\Cref{fig:summary} shows the headline comparison: safety drop and
overrefusal cost for each regime on both models.

\begin{table*}[t]
  \centering
  \caption{Test-set results on Llama-3-8B-Instruct at the
    $50\%$-of-pool checkpoint (\S\ref{sec:cutoff}).
    ASR $=$ attack success rate $(\downarrow)$;
    Ref $=$ refusal rate $(\downarrow)$. Best per column within each
    regime in \textbf{bold}. The two ``adv-benign'' overrefusal sets
    are adversarially-framed benign prompts (look like jailbreaks but
    have benign intent). The WildGuardMix benign set has plain benign
    prompts.}
  \label{tab:test-results}
  \small
  \begin{tabular}{lcccccc}
  \toprule
   & \multicolumn{3}{c}{Safety (ASR $\downarrow$)} & \multicolumn{3}{c}{Overrefusal (Refusal $\downarrow$)} \\
  \cmidrule(lr){2-4} \cmidrule(lr){5-7}
  Baseline & WildJailbreak & WildGuardMix & ClearHarm & \makecell{WildJailbreak\\adv-benign} & \makecell{WildGuardMix\\adv-benign} & \makecell{WildGuardMix\\benign} \\
  \midrule
  \multicolumn{7}{l}{\textit{Pure baselines (trained on the top $50\%$ of the eligible pool)}} \\
  \quad Base & $11.5\%$ & $3.2\%$ & $1.1\%$ & $\mathbf{22.4\%}$ & $\mathbf{20.2\%}$ & $\mathbf{4.5\%}$ \\
  \quad Control & $12.8\%$ & $4.7\%$ & $6.8\%$ & $32.4\%$ & $28.4\%$ & $4.7\%$ \\
  \quad Hard & $\mathbf{2.1\%}$ & $\mathbf{0.3\%}$ & $0.1\%$ & $83.8\%$ & $80.4\%$ & $32.0\%$ \\
  \quad Random & $3.1\%$ & $0.9\%$ & $\mathbf{0.0\%}$ & $78.6\%$ & $74.3\%$ & $35.7\%$ \\
  \midrule
  \multicolumn{7}{l}{\textit{Mixed baselines (the same top $50\%$ interleaved $1{:}1$ with an equal-sized adv-benign set)}} \\
  \quad Base & $11.5\%$ & $3.2\%$ & $1.1\%$ & $\mathbf{22.4\%}$ & $\mathbf{20.2\%}$ & $\mathbf{4.5\%}$ \\
  \quad Control (matched compute) & $13.6\%$ & $5.0\%$ & $8.1\%$ & $30.5\%$ & $24.2\%$ & $5.5\%$ \\
  \quad Hard-Mixed & $\mathbf{5.1\%}$ & $\mathbf{2.9\%}$ & $0.1\%$ & $49.0\%$ & $51.2\%$ & $14.3\%$ \\
  \quad Random-Mixed & $7.9\%$ & $4.4\%$ & $\mathbf{0.0\%}$ & $30.0\%$ & $39.8\%$ & $16.7\%$ \\
  \bottomrule
  \end{tabular}
\end{table*}

\begin{table*}[t]
  \centering
  \caption{Test-set results on Llama-3.2-3B-Instruct at the
    $50\%$-of-pool checkpoint (\S\ref{sec:cutoff}). Best per column
    within each regime in \textbf{bold}. 3B's eligible pool is
    slightly larger than 8B's (\cref{tab:yield}), so $50\%$
    corresponds to a later optimizer step here than in
    \cref{tab:test-results}.}
  \label{tab:test-results-3b}
  \small
  \begin{tabular}{lcccccc}
  \toprule
   & \multicolumn{3}{c}{Safety (ASR $\downarrow$)} & \multicolumn{3}{c}{Overrefusal (Refusal $\downarrow$)} \\
  \cmidrule(lr){2-4} \cmidrule(lr){5-7}
  Baseline & WildJailbreak & WildGuardMix & ClearHarm & \makecell{WildJailbreak\\adv-benign} & \makecell{WildGuardMix\\adv-benign} & \makecell{WildGuardMix\\benign} \\
  \midrule
  \multicolumn{7}{l}{\textit{Pure baselines (trained on the top $50\%$ of the 3B eligible pool)}} \\
  \quad Base & $20.1\%$ & $6.5\%$ & $8.0\%$ & $\mathbf{14.3\%}$ & $\mathbf{21.8\%}$ & $5.1\%$ \\
  \quad Control & $11.8\%$ & $6.2\%$ & $10.6\%$ & $17.6\%$ & $28.1\%$ & $\mathbf{4.1\%}$ \\
  \quad Hard & $1.1\%$ & $\mathbf{0.0\%}$ & $\mathbf{0.0\%}$ & $90.0\%$ & $94.1\%$ & $55.1\%$ \\
  \quad Random & $\mathbf{0.9\%}$ & $\mathbf{0.0\%}$ & $\mathbf{0.0\%}$ & $91.4\%$ & $92.3\%$ & $43.5\%$ \\
  \midrule
  \multicolumn{7}{l}{\textit{Mixed baselines (the same top $50\%$ interleaved $1{:}1$ with an equal-sized adv-benign set)}} \\
  \quad Base & $20.1\%$ & $6.5\%$ & $8.0\%$ & $\mathbf{14.3\%}$ & $\mathbf{21.8\%}$ & $5.1\%$ \\
  \quad Control (matched compute) & $15.7\%$ & $8.2\%$ & $16.6\%$ & $15.2\%$ & $26.8\%$ & $\mathbf{3.9\%}$ \\
  \quad Hard-Mixed & $\mathbf{3.4\%}$ & $\mathbf{1.8\%}$ & $\mathbf{0.0\%}$ & $66.2\%$ & $72.5\%$ & $19.0\%$ \\
  \quad Random-Mixed & $6.8\%$ & $3.8\%$ & $1.7\%$ & $52.4\%$ & $54.5\%$ & $13.7\%$ \\
  \bottomrule
  \end{tabular}
\end{table*}

\subsection{Pure Adversarial Fine-Tuning}
\label{sec:results-pure}

The pure blocks of \cref{tab:test-results,tab:test-results-3b} show
the baselines after each model has trained on the top $50\%$ of its
own eligible pool. Pure
adversarial fine-tuning does what it sets out to on
both models: WildJailbreak ASR drops from $11.5\%$ to
$2.1\text{--}3.1\%$ on 8B and from $20.1\%$ to $0.9\text{--}1.1\%$
on 3B, WildGuardMix drops to under $1\%$ on both ($\leq 0.9\%$ on
8B, $0.0\%$ on 3B), and ClearHarm drops to essentially zero on both.
But it has a large cost in overrefusal. Refusal on the two
jailbreak-shaped benign sets rises from $20\text{--}22\%$ at base to
$74\text{--}84\%$ on 8B, and from $14\text{--}22\%$ to
$90\text{--}94\%$ on 3B, a larger jump for the smaller model.
Refusal on plain benign prompts (the WildGuardMix benign set) rises
from $4.5\%$ to $32\text{--}36\%$ on 8B and from $5.1\%$ to
$44\text{--}55\%$ on 3B.

The compute-matched Control baseline does not explain this
jump. Control rises by $8\text{--}10$~pp on the two adv-benign sets
on 8B and by $3\text{--}6$~pp on 3B, and barely moves on plain
benign ($+0.2$~pp on 8B, $-1$~pp on 3B). The Hard baseline, by
contrast, rises by about $60$~pp on the adv-benign sets on 8B and by
$72\text{--}76$~pp on 3B, at least $6\times$ Control's effect on 8B
and at least $11\times$ on 3B. The overrefusal jump is therefore
caused by the adversarial training data itself, not by generic side
effects of LoRA fine-tuning.

\subsection{$1{:}1$ Adversarially-Framed Benign Mixing}
\label{sec:results-mixed}

The mixed blocks of \cref{tab:test-results,tab:test-results-3b}
show the baselines after training on the same top $50\%$ of hardest
prompts as in the pure regime, in the same order, interleaved $1{:}1$
with the same number of adversarially-framed benign prompts. The adversarial half of mixed
training is identical to the pure regime, only the benign
half is added.

The cost on safety is modest on both models. On WildJailbreak, ASR
rises from $2.1\%$ to $5.1\%$ for Hard-Mixed and from $3.1\%$ to
$7.9\%$ for Random-Mixed on 8B ($+3.0$ and $+4.8$ percentage points),
and from $1.1\%$ to $3.4\%$ and $0.9\%$ to $6.8\%$ on 3B ($+2.3$ and
$+5.9$~pp). The three safety benchmarks still sit well below the
base model on both. The benefit on overrefusal is much larger.
Refusal on the two jailbreak-shaped benign sets drops from
$74\text{--}84\%$ in the pure regime to $30\text{--}51\%$ in the
mixed regime on 8B, and from $90\text{--}94\%$ to $52\text{--}72\%$
on 3B. Refusal on plain benign prompts drops from $32\text{--}36\%$
to $14\text{--}17\%$ on 8B and from $44\text{--}55\%$ to
$14\text{--}19\%$ on 3B.

The size of the overrefusal reduction depends on the baseline.
Random-Mixed cuts adv-benign refusal by $35\text{--}49$~pp (a
$41\text{--}62\%$ relative drop from the corresponding pure
baseline), roughly halving the damage. Hard-Mixed cuts by less:
$22\text{--}35$~pp ($23\text{--}42\%$ relative drop), about a third
of the damage. On plain benign, both Hard-Mixed and Random-Mixed
reduce refusal by more than half ($53\text{--}69\%$ relative).

\subsection{Hardest Half vs Random Half}
\label{sec:results-curriculum}

On safety, training on the hardest half helps clearly in the mixed
regime on both models. The gap on WildJailbreak ASR is $2.8$~pp on
8B (Hard-Mixed $5.1\%$ vs Random-Mixed $7.9\%$) and $3.4$~pp on 3B
($3.4\%$ vs $6.8\%$), similar magnitudes. In the pure regime, the
gap is smaller and not always in Hard's favor: $1.0$~pp on 8B
($2.1\%$ vs $3.1\%$) and $-0.2$~pp on 3B ($1.1\%$ vs $0.9\%$). The
3B pure case is within noise: Hard and Random both drive 3B's
pure-baseline ASR essentially to the floor on all three safety sets
(\cref{tab:test-results-3b}), so there is little room for prompt
selection to make a difference. In the mixed regime, ASR stays
above the floor on both models, so the selection effect becomes
visible.

On refusal of jailbreak-shaped benign prompts, the comparison
flips: in the mixed regime, Random-Mixed beats Hard-Mixed on both
models. On 8B, Random-Mixed achieves $30.0\%$ and $39.8\%$ on the
two adv-benign sets vs Hard-Mixed's $49.0\%$ and $51.2\%$ (an
$11\text{--}19$~pp gap in Random's favor). On 3B, the gap is
$13.8$~pp ($52.4\%$ vs $66.2\%$ on WildJailbreak adv-benign) and
$18.0$~pp ($54.5\%$ vs $72.5\%$ on WildGuardMix adv-benign).
A possible explanation is that training on only the hardest
prompts pushes the model toward aggressive refusal on
jailbreak-shaped inputs, while Random's broader half (which also
includes easier-looking adversarial prompts) teaches finer
discrimination that generalizes to adv-benign. Neither pure regime
shows this flip.

On plain benign prompts, the gap is small and inconsistent across
models, so the mechanism seems tied to jailbreak-shaped benign
prompts specifically.

\section{Discussion}

\paragraph{Why adv-benign rather than vanilla benign in the mix?}
Vanilla benign prompts look benign on the surface, and that is not
where overrefusal fails. Adversarially-framed benign prompts have
the same surface structure as jailbreaks (role-play, hypothetical,
persona preambles) but ask for harmless content. Mixing these in is
what teaches the model to tell ``looks like a jailbreak'' apart from
``actually asks for harm'' on the failure-mode distribution.
The residual adv-benign refusal in the best mixed baseline is
therefore the right figure to report: it is the hard part of the
problem, not the easy one. The gap between the two models suggests
that model capacity also matters for this discrimination task,
even when the training signal is right.

\paragraph{Pure baselines reduce ASR more. Is that the right
comparison?} No. A model that refuses every prompt also has zero
ASR, but it has zero utility. The right comparison is Pareto over
(ASR, overrefusal): on both models, the mixed baselines sit at
strictly better (ASR, overrefusal) tradeoffs than any pure baseline
on both adv-benign sets, at a small ASR cost.

\paragraph{Focusing on the hardest prompts is a tradeoff.} The
hardest-half-vs-random-half comparison
(\S\ref{sec:results-curriculum}) cuts in opposite directions across
the two failure modes: focusing on the hardest prompts protects
safety better, while the broader Random half generalizes
better on adv-benign refusal. Hardest-first is therefore not
strictly preferable. It shifts the (ASR, overrefusal) tradeoff
toward safety.

\section{Limitations and Future Work}

\textbf{Limitations.} All reported results are on the Llama-3
family (Llama-3-8B-Instruct and Llama-3.2-3B-Instruct), so
cross-family generalization remains untested. Even the best mixed
baseline refuses $30\text{--}51\%$ (8B) or $52\text{--}72\%$ (3B)
of jailbreak-shaped benign prompts. Mixing reduces the damage but
does not close the gap to base. The $50\%$-of-pool cutoff was
chosen after inspecting the hardness distributions, and although
we argue in \S\ref{sec:cutoff} that the justification depends only
on the shape of the difficulty distribution, the post-hoc framing
remains a caveat.

\textbf{Future work.}
Future work should refine both the definition and use of prompt hardness. Our current hardness score is model-conditioned: a prompt is considered hard when the target model produces harmful responses on a large fraction of sampled rollouts. This is useful for mining model-specific failures, but does not show whether the prompt is intrinsically hard. Hardness could also be extended beyond frequency by incorporating the severity and semantic diversity of harmful rollouts, since prompts that elicit many distinct harmful completions may expose broader safety failures than prompts that repeatedly elicit the same response. Another direction is to use the always-jailbroken prompts currently dropped by our pipeline: a stronger safety-aligned teacher could generate a safe response, the target model could rewrite it into its own distribution, and the result could be filtered by the same judges. Finally, future work should study curricula that explicitly balance ASR, overrefusal, and training efficiency, such as starting with adversarially-framed benign prompts before gradually increasing high-hardness harmful prompts.

\section{Conclusion}

We described a self-contained safety fine-tuning pipeline in which
the adversarial training data and the supervised targets both come
from the target model's own rollouts. Five controlled baselines
built on a canonical prompt-to-target pairing separate the effects
of prompt selection and benign interleaving. On Llama-3-8B-Instruct
and Llama-3.2-3B-Instruct, pure adversarial fine-tuning produces a
clear safety-vs-overrefusal tradeoff that compute-matched controls
do not explain, and $1{:}1$ adversarially-framed benign mixing
recovers a meaningful share of the lost compliance at a small
safety cost ($2\text{--}6$~pp ASR on WildJailbreak). The
hardest-half-vs-random-half comparison points in opposite directions
across safety and adv-benign refusal, so prompt selection shifts
the (ASR, overrefusal) tradeoff rather than dominating it. The
smaller model retains higher residual refusal even after mixing,
suggesting model capacity also matters.

\bibliography{example_paper}
\bibliographystyle{icml2026}

\newpage
\appendix
\onecolumn

\section{Baseline Construction Procedure}
\label{app:algo}

\Cref{alg:mining} is the construction procedure referenced in
\S\ref{sec:baselines}.

\begin{algorithm}[H]
  \caption{Self-Mined Baseline Construction}
  \label{alg:mining}
  \begin{algorithmic}[1]
    \STATE {\bfseries Input:} target model $M$; adversarial-harmful prompts
    $\mathcal{P}_h$; adversarial-benign prompts $\mathcal{P}_b$;
    vanilla-benign prompts $\mathcal{V}$; sample sizes $N_h, N_b$;
    rollout counts $K_h, K_b$; safety judges $J_1, J_2, J_3$; refusal
    judge $J_r$; seed.
    \STATE Sample $\mathcal{P}_h \leftarrow$ uniform $N_h$-subset of
    \texttt{adversarial\_harmful}.
    \STATE Sample $\mathcal{P}_b \leftarrow$ uniform $N_b$-subset of
    \texttt{adversarial\_benign}.
    \STATE \COMMENT{Hard-example mining}
    \FOR{$p \in \mathcal{P}_h$}
      \STATE Draw $K_h$ rollouts $r_{p,1}, \dots, r_{p,K_h} \sim M(\cdot \mid p)$ at $T = 1$.
      \STATE Compute $\overline{J}(p, r_{p,i})$ via majority vote of $J_1, J_2, J_3$.
      \STATE Set $\mathrm{hr}(p), \mathrm{ur}(p)$ as in \S\ref{sec:mining}.
    \ENDFOR
    \STATE $\mathcal{E} \leftarrow \{p : 0 < \mathrm{hr}(p) < 1\}$; $n \leftarrow |\mathcal{E}|$.
    \STATE \COMMENT{Canonical safe-target map}
    \FOR{$p \in \mathcal{E}$}
      \STATE $\sigma(p) \leftarrow$ uniform draw from $\{i \mid \overline{J}(p, r_{p,i}) = \textsc{unharmful}\}$.
      \STATE $\mathrm{safe\_target}(p) \leftarrow r_{p,\sigma(p)}$.
    \ENDFOR
    \STATE \COMMENT{Adversarial-benign mining}
    \FOR{$q \in \mathcal{P}_b$}
      \STATE Draw $K_b$ rollouts; classify each with $J_r$.
    \ENDFOR
    \STATE $\mathcal{E}_b \leftarrow \{q : \exists\, i,\, J_r(q, r_{q,i}) = \textsc{complied}\}$.
    \STATE $\mathcal{B} \leftarrow$ uniform $n$-subset of $\mathcal{E}_b$; draw canonical $\mathrm{compliant\_target}(q)$ from non-refused rollouts.
    \STATE \COMMENT{Baselines}
    \STATE $\mathcal{D}_{\text{hard}} \leftarrow \mathcal{E}$ sorted stably by $\mathrm{hr}$ descending.
    \STATE $\mathcal{D}_{\text{random}} \leftarrow$ random permutation of $\mathcal{E}$.
    \STATE $\mathcal{D}_{\text{control}} \leftarrow n$ prompts from $\mathcal{V}$; one $T{=}1$ rollout each.
    \STATE $\mathcal{D}_{\text{Hard-Mixed}} \leftarrow \textsc{interleave}(\mathcal{D}_{\text{hard}}, \mathcal{B})$.
    \STATE $\mathcal{D}_{\text{Random-Mixed}} \leftarrow \textsc{interleave}(\mathcal{D}_{\text{random}}, \mathcal{B})$.
    \STATE {\bfseries Output:} five training JSONL files (preserved order).
  \end{algorithmic}
\end{algorithm}

\section{Training Hyperparameters}
\label{app:hyperparams}

\begin{table}[H]
  \caption{LoRA training hyperparameters.}
  \label{tab:hyperparams}
  \vskip 0.1in
  \begin{center}
    \begin{small}
      \begin{tabular}{ll}
        \toprule
        Hyperparameter                       & Value             \\
        \midrule
        LoRA rank $r$                        & $16$              \\
        LoRA scaling $\alpha$                & $32$              \\
        LoRA dropout                         & $0.05$            \\
        LoRA target modules                  & $\{q\_proj, v\_proj\}$ \\
        Per-device batch size                & $2$               \\
        Gradient accumulation                & $5$               \\
        Effective batch size                 & $10$              \\
        Epochs                               & $1$               \\
        Learning rate                        & $1 \times 10^{-4}$ \\
        Schedule                             & cosine, warmup ratio $0.03$ \\
        Weight decay                         & $0.01$            \\
        Optimizer                            & AdamW             \\
        Precision                            & bf16              \\
        Gradient checkpointing               & on (non-reentrant) \\
        Max sequence length                  & $1024$            \\
        Sampler                              & sequential (no shuffle) \\
        Checkpoint interval                  & every $5$ optimizer steps ($50$ prompts) \\
        Seed                                 & $55$              \\
        \bottomrule
      \end{tabular}
    \end{small}
  \end{center}
  \vskip -0.1in
\end{table}

\section{Hardness Distribution on the Eligible Pool}
\label{app:hardness}

\Cref{tab:hardness-dist} gives the full harmful-rate distribution
over the eligible pool for both target models. Buckets are
left-inclusive, right-exclusive: the $10\%$ boundary appears only
in the $10\text{--}20\%$ row. Both distributions are heavily
right-skewed. Together with \cref{tab:yield}, this is the
quantitative basis for the $50\%$-of-pool cutoff argued in
\S\ref{sec:cutoff}. 3B sits slightly higher in the mid-bands
($40$--$79\%$) and slightly lower at the floor, consistent with
its higher base ASR ($20.1\%$ vs $11.5\%$).

\begin{table}[H]
  \caption{Distribution of harmful rate $\mathrm{hr}(p)$ over the
    eligible pool $\mathcal{E}$ on both target models (counts and
    percentages of $\mathcal{E}$). Buckets are left-inclusive,
    right-exclusive.}
  \label{tab:hardness-dist}
  \vskip 0.1in
  \begin{center}
    \begin{small}
      \begin{tabular}{lrrrr}
        \toprule
        $\mathrm{hr}(p)$ bucket & 8B count & 8B \% & 3B count & 3B \% \\
        \midrule
        $\geq 80\%$       & $115$ & $4.8\%$  & $116$ & $4.7\%$  \\
        $60\text{--}80\%$ & $124$ & $5.2\%$  & $174$ & $7.0\%$  \\
        $40\text{--}60\%$ & $197$ & $8.2\%$  & $267$ & $10.7\%$ \\
        $20\text{--}40\%$ & $370$ & $15.5\%$ & $406$ & $16.3\%$ \\
        $10\text{--}20\%$ & $339$ & $14.2\%$ & $360$ & $14.5\%$ \\
        $5\text{--}10\%$  & $502$ & $21.0\%$ & $465$ & $18.7\%$ \\
        $< 5\%$           & $741$ & $31.0\%$ & $700$ & $28.1\%$ \\
        \midrule
        Total $|\mathcal{E}|$ & $\mathbf{2{,}388}$ & & $\mathbf{2{,}488}$ & \\
        \bottomrule
      \end{tabular}
    \end{small}
  \end{center}
  \vskip -0.1in
\end{table}

\section{LLM Usage}
\label{app:llm}

We used LLMs (ChatGPT and Claude) to assist with writing edits,
\LaTeX{} formatting, and code. All technical content,
experiments, and claims were produced and verified by the authors.

\end{document}